\documentclass{article}

\usepackage{times}
\usepackage{graphicx} % more modern
\usepackage{subfigure} 

\usepackage{natbib}

\usepackage{algorithm}
\usepackage{algorithmic}

\usepackage{hyperref}

\usepackage[arXiv]{icml2017}

\usepackage{amsmath}

\icmltitlerunning{A Cascaded CNN for X-ray Low-dose CT Image Denoising}

\begin{document} 

\twocolumn[
\icmltitle{A Cascaded Convolutional Nerual Network \\ 
		for X-ray Low-dose CT Image Denoising}



% you can specify symbols, otherwise they are numbered in order
\icmlsetsymbol{equal}{*}

\begin{icmlauthorlist}
\icmlauthor{Dufan Wu}{to}
\icmlauthor{Kyungsang Kim}{to}
\icmlauthor{Georges El Fakhri}{to}
\icmlauthor{Quanzheng Li}{to}
\end{icmlauthorlist}

\icmlaffiliation{to}{Gordon Center for Medical Imaging, Massachusetts General Hospital and Harvard Medical School, Boston, MA, USA}

\icmlcorrespondingauthor{Quanzheng Li}{li.quanzheng@mgh.harvard.edu}

\icmlkeywords{computed tomography, low-dose, image denoising, machine learning, artificial neural network}

\vskip 0.3in
]



\printAffiliationsAndNotice{}  % leave blank if no need to mention equal contribution
%\footnotetext{hi}

\begin{abstract} 
Image denoising techniques are essential to reducing noise levels and enhancing diagnosis reliability in low-dose computed tomography (CT). Machine learning based denoising methods have shown great potential in removing the complex and spatial-variant noises in CT images. However, some residue artifacts would appear in the denoised image due to complexity of noises. A cascaded training network was proposed in this work, where the trained CNN was applied on the training dataset to initiate new trainings and remove artifacts induced by denoising. A cascades of convolutional neural networks (CNN) were built iteratively to achieve better performance with simple CNN structures. Experiments were carried out on 2016 Low-dose CT Grand Challenge datasets to evaluate the method's performance. 
\end{abstract} 

\section{Introduction}
\label{introduction}

X-ray computed tomography (CT) is an essential method for non-invasive diagnosis in modern medicine. To reduce patients' radiation dose and related risk of cancers, low-dose CT has been of great research interest in CT imaging. Due to reduced number of X-ray photons in low-dose CT, there is significantly increased noise levels and artifacts in the image, which decreases the reliability of diagnosis \cite{kalra2004strategies}. Thus, image denoising methods are necessary for low-dose CT imaging. However, since the noises in CT images are non-Gaussian, spatial-variant and related to surrounding structures \cite{hsieh2003computed}, it is very hard for conventional denoising algorithms to handle. 

By learning nonlinear filter banks to map noisy images to less noisy ones in the training dataset, neural network based denoising methods were able to handle complicated noises. Jain and Seung demonstrated that a convolutional neural network (CNN) based denoiser could achieve better peak signal to noise ratio (PSNR) compared to conventional denoisers for additive white Gaussian noise \yrcite{jain2009natural}. Burger et al. used multilayer perceptron (MLP) for natural image denoising, and achieved much better performance on non-Gaussian noises such as pepper and salt noises and JPEG compression artifacts compared to BM3D and other conventional denoisers \yrcite{burger2012image}. The neural networks were further extended to removing more complex corruptions of images, such as inpainting, deconvolution, removal of rain streaks, droplets and dirt, etc \cite{xie2012image, agostinelli2013adaptive, xu2014deep, fu2016clearing, eigen2013restoring}. 

The denoising of low-dose CT has been challenging for decades. The noises roughly follow a mixed distribution of Poisson and Gaussian with spatial-variant deviations in the Radon transform domain of the CT image \cite{elbakri2002statistical}, and it becomes highly correlated to its surrounding structures in the image domain. On the other hand, CT based diagnosis such as liver lesion detection required high resolution and contrast, thus any blur or artifacts generated by denoising might cause confusion for clinical diagnosis \cite{prakash2010reducing, pickhardt2012abdominal}. Some recent works showed promising results by applying CNN based denoisers to low-dose CT images \cite{kang2016deep, chen2017low}. The CNN was trained from normal-dose images and their corresponding low-dose images which were generated by adding noises in the Radon transform domain according to physical models. They achieved much better performance, especially for low-contrast details such as vessels and lesions.

However, in our studies it was discovered that sometimes the denoised images contain unnatural artifacts induced by the CNN denoisers, mainly due to the high complexity of noise and lack of sufficient training data. Rather than initiating completely new training with deeper CNNs, we proposed a cascaded framework to boost performance of simple CNNs. The low-dose images in the training dataset was denoised with the trained CNN, then a second level CNN was trained to map the denoised results to normal-dose images, where the artifacts generated by the first level CNN could be reduced. This scheme could be executed recurrently to get deeper cascades and further refine the denoised images. Experiments on 2016 Low-dose CT Grand Challenge \cite{aapm2016challenge} demonstrated that the proposed method could effectively reduce artifacts in CT images that were denoised with CNN.  

\section{Method}
\subsection{CNN Denoiser}

The simple residue learning CNN proposed by Zhang et al. was used in this work \yrcite{zhang2016beyond}, but other more complex neural networks such as RED-Net can also fit into the proposed training framework easily \cite{mao2016image}. The CNN mapped the low-dose images to the difference between low- and normal-dose images rather than the normal-dose image itself, which gave a better conditioned problem for the training \cite{han2016deep}. 

\begin{figure}[t]
\vskip 0.2in
\begin{center}
\centerline{\includegraphics[width=\columnwidth]{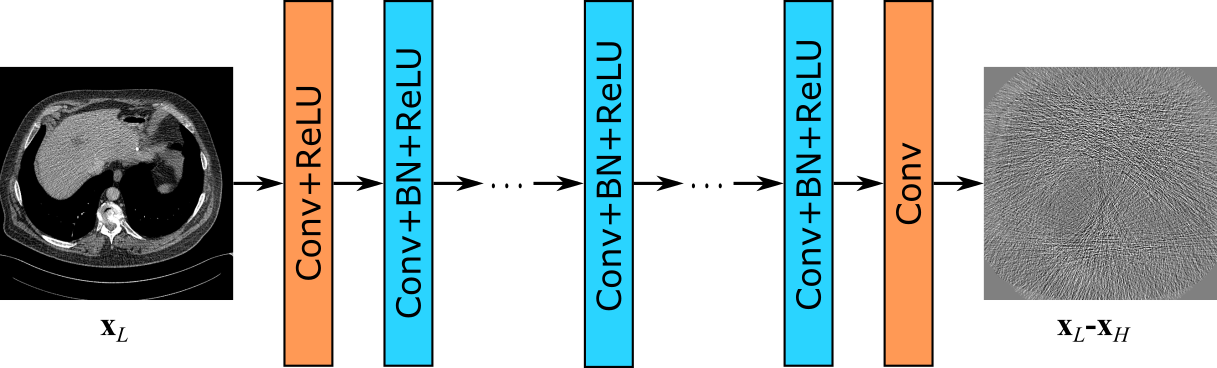}}
\caption{The structure of the CNN denoiser. Conv stands for convolution layers, ReLU stands for ReLU layers, and BN stands for batch normalization.}
\label{fig:CNN}
\end{center}
\vskip -0.2in
\end{figure}

The structure of the used CNN is shown in Figure~\ref{fig:CNN}. It consisted of sequential convolution modules which were composed of convolution, batch normalization \cite{ioffe2015batch} and ReLU layers, except that the first module excluded batch normalization layer and the last module contained only convolution layer. $3\times3\times64$ kernels with zero padding were used for all the convolution layers except that the output layer only had one feature map. L2 loss function was used during the training.

\subsection{Cascaded Training}

Due to limited capacity of CNN, amount of data and the training time, the CNN was not guaranteed to remove all the noises. Furthermore, sometimes the denoising would induce new artifacts into the image when it encountered noises which were too strong or had patterns that were rarely seen in the training dataset. Conventional solutions includes improvement on CNN's structures and depth and feeding CNN with more training samples, where each training were independent and previous trained networks could not be used.

However, despite of the residue artifacts in the results, the trained CNN denoisers still suppressed most of the noises. Thus, the CNN could be applied to the low-dose images in the training dataset for initial noise reduction. Then the new CNN trained on the processed training dataset would focus on the residue artifacts to remove them. The process could be done iteratively and build a cascade of CNNs which refine the denoising results gradually. 

\begin{figure}[t]
\vskip 0.2in
\begin{center}
\centerline{\includegraphics[width=\columnwidth]{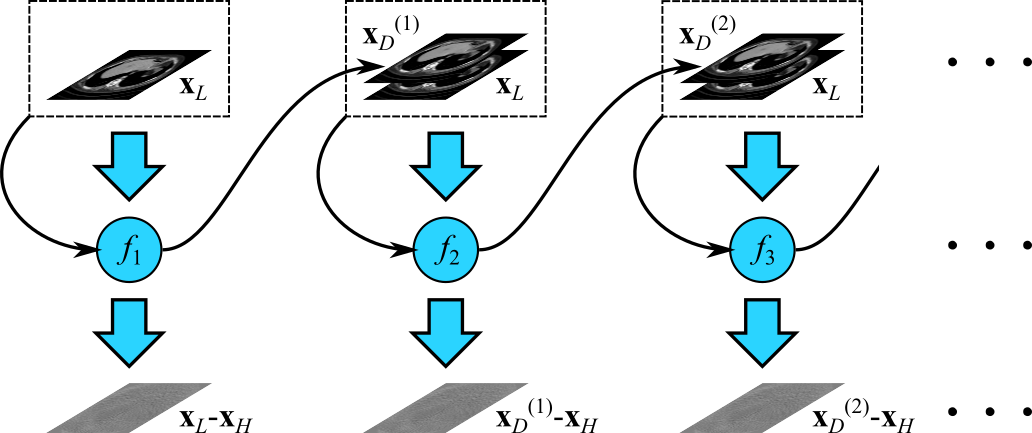}}
\caption{The cascaded CNN training scheme. The big blue arrows stand for CNN training, and the narrow black arrows stand for denoising with the trained CNN. $f_k$ is the CNN at the {\it k}th cascade. }
\label{fig:cascade}
\end{center}
\vskip -0.2in
\end{figure}

The training scheme of the cascaded CNN is shown in Figure~\ref{fig:cascade}. Denote the low- and normal-dose images in the training dataset as $\mathbf{x}_L$ and $\mathbf{x}_H$. A denoising CNN $f_1$ with the structure in Figure~\ref{fig:CNN} was firstly trained to map $\mathbf{x}_L$ to the residue $\mathbf{x}_H$. Then $\mathbf{x}_L$ was denoised with $f_1$ to get the denoised images $\mathbf{x}_D^{(1)}$. Then a cascaded CNN $f_2$ was trained to map $\mathbf{x}_D^{(1)}$ to $\mathbf{x}_D^{(1)}-\mathbf{x}_H$ in a similar fashion. However, to prevent information loss, the inputs to $f_2$ and following cascaded CNNs were constructed by stacking $\mathbf{x}_L$ and $\mathbf{x}_D^{(1)}$ across channels. The following cascaded were constructed in the same way with the previous one, as demonstrated in Figure~\ref{fig:cascade}. All the cascaded CNNs had the same structure except for the number of input channels. It should be noted that although training was completed on patches to save memories during training, the denoising performed on the training dataset was on the entire image so that training patches would not get artifacts near the boundaries. 

\subsection{Relation to Existing Works}

\begin{figure}[t]
\vskip 0.2in
\begin{center}
\centerline{\includegraphics[width=\columnwidth]{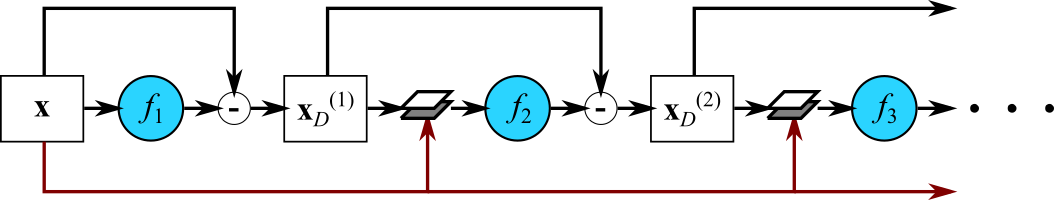}}
\caption{The structure of the cascaded CNN in the test phase. The red line indicates the different between it and a residue network.}
\label{fig:inference}
\end{center}
\vskip -0.2in
\end{figure}

As demonstrated in Figure~\ref{fig:inference}, the structure of the cascaded CNN resembled that of a residue neural network \cite{he2016deep}. The main difference was that in a residue network, there are bypasses to add the feature maps in different layers, but the bypasses in the proposed cascaded CNN was to stack the original input with the intermediate results. The bypasses preserved information and allowed us to partition a deep network into small parts and train them step by step. 

The iterative training scheme also resembled the sparse stacked denoising auto-encoders (SSDA) for image denoising \cite{xie2012image} (which worked differently from the stacked denoising auto-encoders for feature extraction \cite{vincent2008extracting}). In SSDA, the network were trained from outermost to innermost and one layer at a time to make the problem easier to solve. But it required a finalizing step to optimize the entire network in the end. The proposed cascaded CNN did not require such a step. 

\section{Experimental Setup}

\subsection{Datasets}

The data from 2016 Low-dose CT Grand Challenge \cite{aapm2016challenge} were used for the experiments. It contained abdominal CT images collected from 10 different patients with Siemens CT scanners with 120kVP tube voltage and 200 effective mAs dose. The low-dose images were simulated by adding Poisson noises to each patient's projection data and the amplitude of noises were equivalent to quarter-dose acquisition. The low- and normal-dose images were both reconstructed from the projection data with filtered backprojection (FBP) algorithm on the scanners. The axial resolution of the images were $512\times512$, but different patients may have different pixel size between $0.66\text{mm}\times0.66\text{mm}$ to $0.8\text{mm}\times0.8\text{mm}$. Each patient had multiple slices of images and slice thickness was $1\text{mm}$ for all the patients. 

\subsection{Training Parameters}

Images from 7 patients were used as the training datasets and the CNN denoisers were trained on 2D slices. There were 3,933 slices in total in the training dataset, and 150 patches with size of $40\times40$ were randomly extracted from each slice at each cascade. The order of the patches were shuffled within each patient but there was no inter-patient shuffle. The patches were divided by a factor of 512 Housefield units (HU) for normalization. 

The convolution kernels in a single denoising CNN was $3\times3$ with an output feature map with 64 channels except for the last layer, where there was only 1 channel. Zero padding was used to keep the input and output of the same size. The weights of the convolution kernels were initialized with Xavier weightings and 0 biases \cite{glorot2010understanding}. 3 different depths of CNN were used in the study, where the number of "Conv+BN+ReLU" modules in Figure~\ref{fig:CNN} were 5, 10, and 15. The CNNs were denoted as CNN5, CNN10, and CNN15 respectively. 

The CNN in each cascade were trained with Adaptive Moment Estimation (ADAM) method \cite{kingma2014adam} with a minibatch of 100 patches. It was trained with a weight penalty of $10^{-4}$. The parameters for ADAM was chosen as suggestions in the original work, except that the learning rate was set to $10^{-4}$. A total iteration number of 90,000 was used for training, which was equivalent to approximately 15 epoches on the training dataset. The neural network were implemented with Caffe \cite{jia2014caffe}. 

\subsection{Comparison Methods}

Conventional image denoising methods, block-matching and 3D filtering (BM3D) and weighted nuclear norm minimization (WNNM) \cite{dabov2007image, gu2014weighted}, were used for comparison study. A machine learning based method, multilayer perceptron (MLP) was also used for analysis of the cascaded training scheme \cite{burger2012image}. 

In BM3D and WNNM, most of the parameters were inherited from the codes provided by the authors\footnote{BM3D code: http://www.cs.tut.fi/~foi/GCF-BM3D/} \footnote{WNNM code: http://www4.comp.polyu.edu.hk/~cslzhang/}, except for the noise level which were tunned for best visualization in the denoised image. For BM3D, the noise level was set to 8\% of the slice's dynamic range, and for WNNM it was set to 4\%. 

Residue learning was not used for MLP, which meant that the trained neural networks directly mapped the low-dose images to normal-dose images instead of the difference between them. The MLP was a fully connected neural network with 2 hidden layer and each of them had 511 nodes. It was trained on $13\times13$ patches. 500 patches were extracted from each slice in the training dataset and normalized and shuffled in the same way with the cascaded CNN. ADAM algorithm with the same parameters were used to train the MLPs with 900,000 iterations. During denoising, $13\times13$ patches with a stride of 3 were extracted from the low-dose images. The denoised patches were aggregated into an entire image with a Gaussian weighting window of $13\times13$ size and standard deviation of 13/3. A cascade was also built with MLPs.

\section{Results}

\subsection{Quantitative Image Quality Estimations}

300 slices that contained liver were extracted from one of the test patients' data and denoised with the trained CNNs. 8, 4, and 4 cascades were built for CNN5, CNN10 and CNN15 respectively. A 4 cascades of MLPs were also trained. The images were denoised slice by slice. Peak to noise ratios (PSNR) were calculated between the denoised images and the corresponding normal-dose images at each cascade, where the dynamic range of the images were considered as 12 bits. Structural similarities (SSIM) \cite{wang2004image} were also calculated in a grayscale window of $[-160\text{HU}, 240\text{HU}]$, which was a common window for liver diagnosis. The SSIMs were calculated slice by slice and averaged together. 

Loss of textures were observed for the denoised images, which made the denoised images look unnatural compared to clinical normal-dose CT images. One of the most widely used method was to blend the denoised images and the low-dose images linearly. We used 70\% denoised images plus 30\% low-dose images to produce the blended results, which gave a similar noise level compared to normal-dose images in the experiments. The blended results would give results that looked more similar to the normal-dose images, which was not noise free. Quantitative measurements including PSNR and SSIM were also calculated for the blended results. 

\begin{figure}[t]
\vskip 0.2in
\begin{center}
\centerline{\includegraphics[width=\columnwidth]{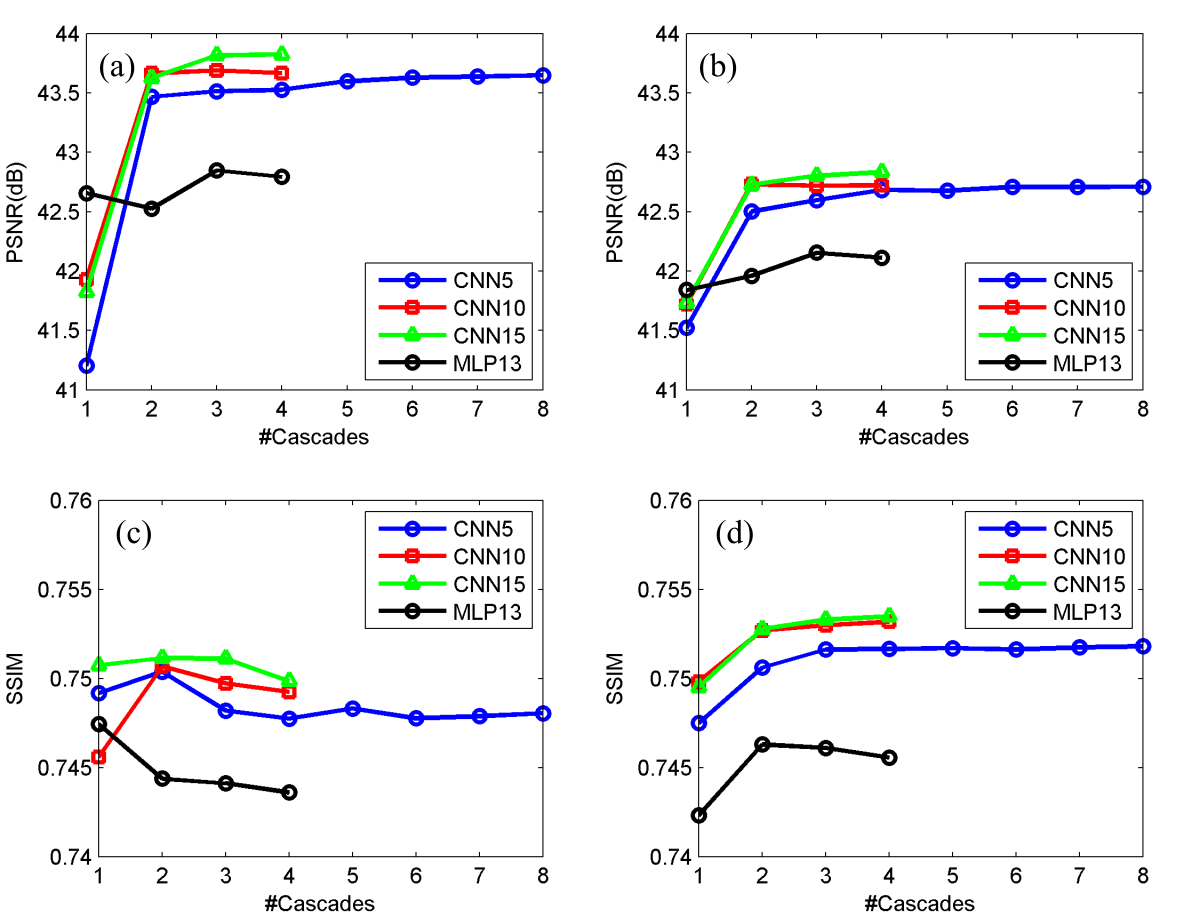}}
\caption{Image quality indices with different neural networks and number of cascades: (a) PSNR of original denoised images; (b) PSNR of blended images; (c) SSIM of original denoised images; (d) SSIM of blended images. Number of cascades equals to the total number of CNNs in the entire denoising chain.}
\label{fig:psnr}
\end{center}
\vskip -0.2in
\end{figure} 

The quantitative results are presented in Figure~\ref{fig:psnr}. For the CNNs, applying the cascading gave a great boost to PSNR for both denoised and blended results. There were some gain on SSIM by applying the first cascade but it deteriorated in the following cascades for original denoised images. However, after blending, the SSIMs increased compared to original denoised images for most of the cases, and the image quality had a increasing trend when applying more cascades. The main reason for such big difference on SSIM for denoised and blended images was that normal-dose images were not noise free in either training dataset or testing dataset. The denoisers would remove most of the noises and gave smooth images, which were actually quite different from the normal-dose images, thus it would produce the unpredictable trend in the SSIMs of original denoised results. 

For the results of MLPs, the gain of cascades were not as good as that of the CNNs. One possible reason was that the power of MLP as a denoiser was not as much as CNN, and the denoised images contained relatively large amount of noises and denoising artifacts, and the artifacts induced by the cascaded MLPs were no less than the artifacts they removed. Although it was not as ideal as the CNN cascades, a significant increase in both PSNR and SSIM when adding the first cascade were observed for the MLPs. Furthermore, the boosts on CNNs' performance were also the largest when the number of cascade went from 1 to 2, which indicated that using one additional training could be effective and robust for denoising applications. 

\subsection{Subjective Image Quality Estimations}

\begin{figure}[t]
\vskip 0.2in
\begin{center}
\centerline{\includegraphics[width=\columnwidth]{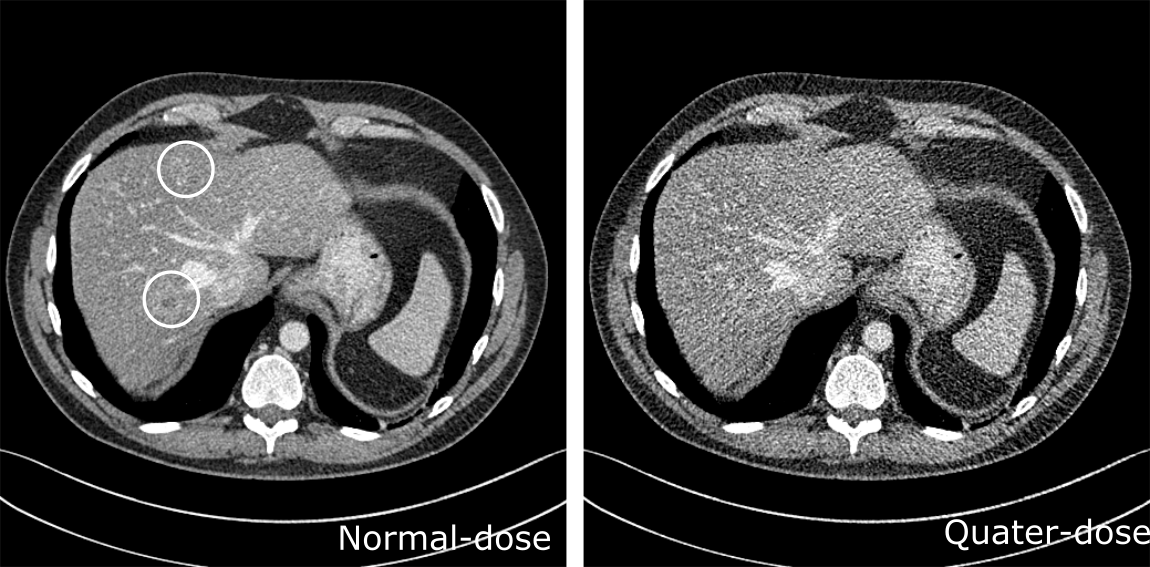}}
\caption{The normal- and low-dose images from the scanner of the selected slice. Metastasis were marked with white circles in the normal-dose image. The grayscale window for display is $[-160\text{HU}, 240\text{HU}]$.}
\label{fig:groundTruth}
\end{center}
\vskip -0.2in
\end{figure} 

\begin{figure*}[t]
\vskip 0.2in
\begin{center}
\centerline{\includegraphics[width=\textwidth]{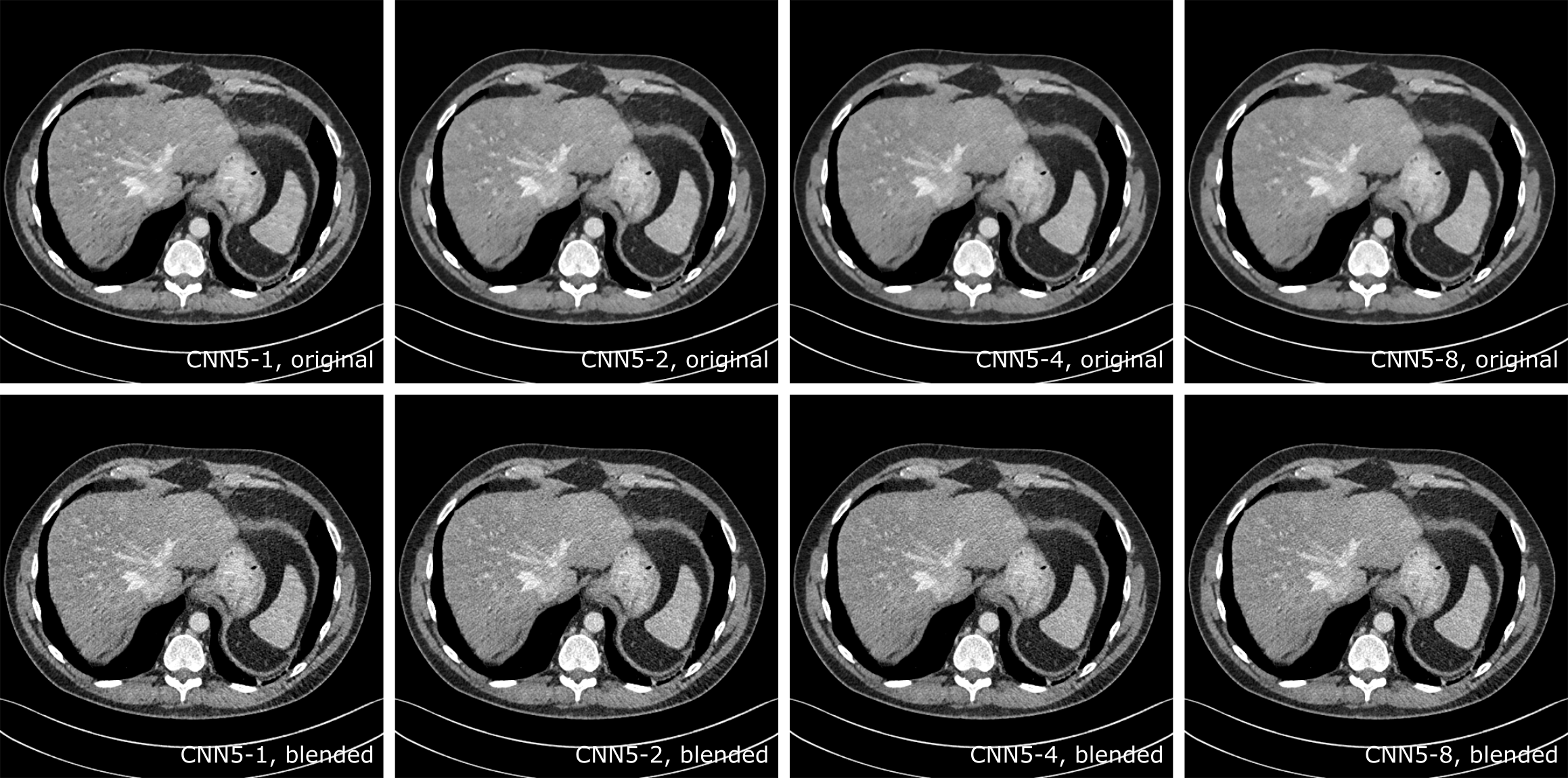}}
\caption{The original denoised and blended image for different cascades of CNN5. CNN5-{\it k} denotes the CNN denoiser with $k$ cascades. The grayscale window for display is $[-160\text{HU}, 240\text{HU}]$.}
\label{fig:CNN5}
\end{center}
\vskip -0.2in
\end{figure*} 

A slice that contained metastasis is selected to demonstrate the image quality improvement with cascaded CNNs visually. Figure~\ref{fig:groundTruth} showed the corresponding normal-dose and original low-dose image of the selected slice, where the two metastasis contained in the slice were marked with white circles. The denoised results and corresponding 70\% blended images are shown in Figure~\ref{fig:CNN5}. Compared to the low-dose image, the visibility of the metastasis, especially the one near the chest, is greatly improved. There were noticeable blocky artifacts in CNN5-1 results in the liver, but it was greatly reduced by applying CNN cascades and almost gone in the results of CNN5-8. For the blended results, the improvement was not obvious as that on the original ones because of the increased noise level. Compared to CNN5-1, the cascaded CNNs gave more uniform liver areas with less speckles. No useful information lost such as blurring was observed with increased number of cascades. 

\begin{figure*}[p]
\vskip 0.2in
\begin{center}
\centerline{\includegraphics[width=\textwidth]{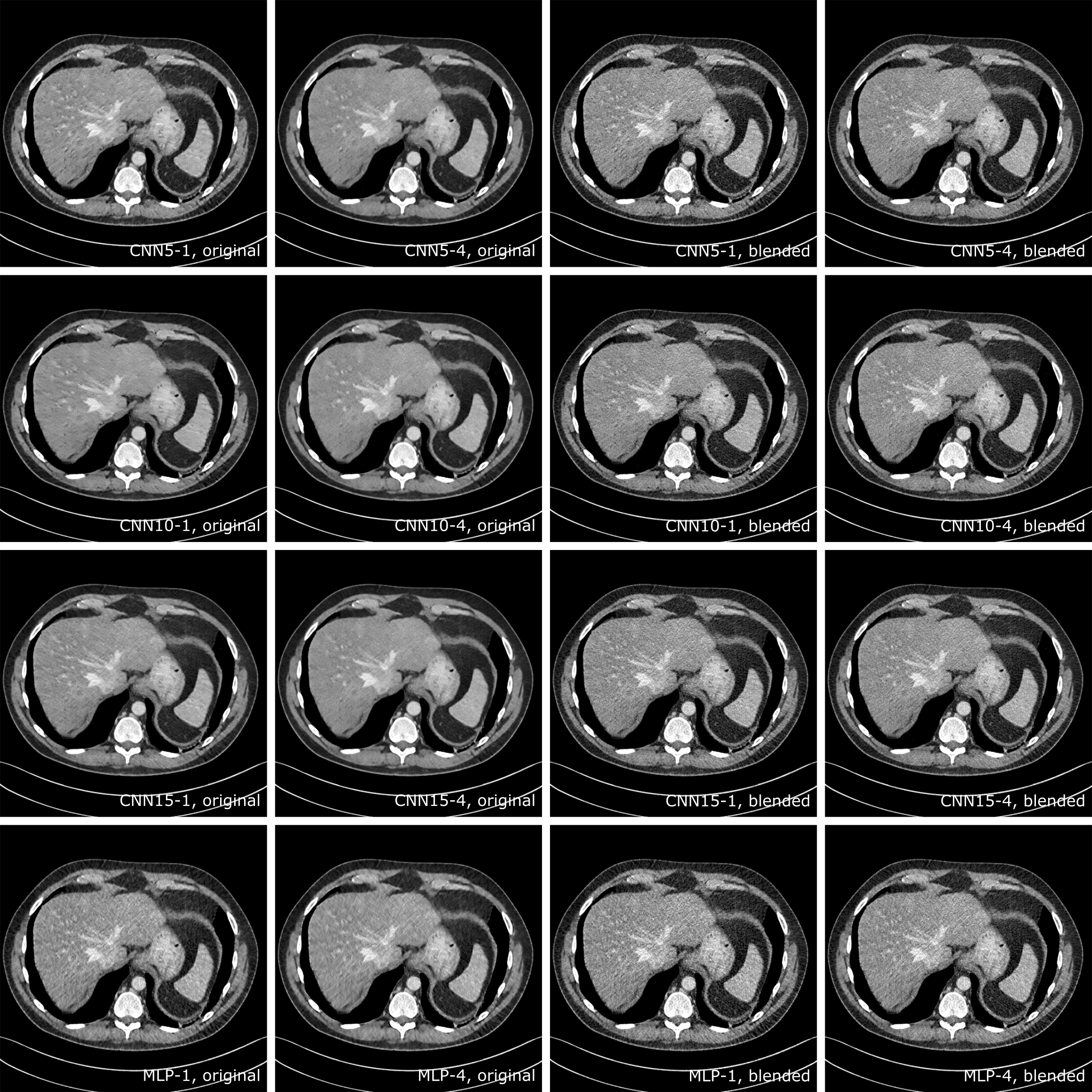}}
\caption{The results produced with different type of neural networks with cascaded training. The grayscale window for display is $[-160\text{HU}, 240\text{HU}]$.}
\label{fig:CNNComp}
\end{center}
\vskip -0.2in
\end{figure*}

\begin{figure*}[t]
\vskip 0.2in
\begin{center}
\centerline{\includegraphics[width=\textwidth]{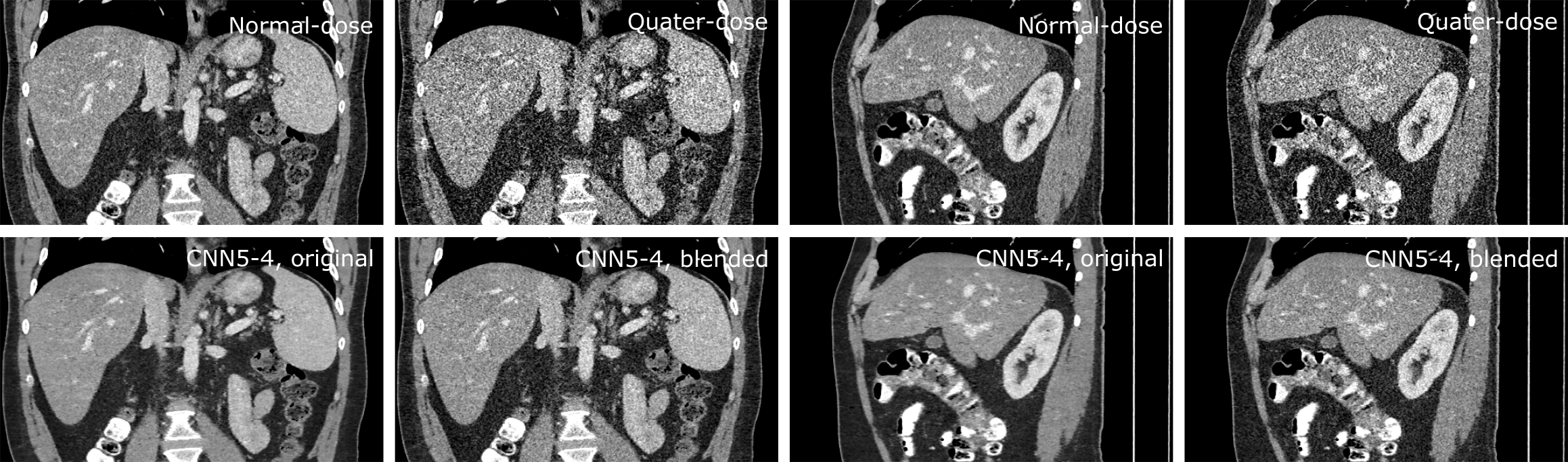}}
\caption{The coronal and sagittal views of the 300 slices that were denoised. The grayscale window for display is $[-160\text{HU}, 240\text{HU}]$.}
\label{fig:SagCor}
\end{center}
\vskip -0.2in
\end{figure*}

Figure~\ref{fig:CNNComp} demonstrated the result's of cascaded CNN5, CNN10, CNN15 and MLP with the cascaded training scheme. There were strip artifacts on the liver for CNN10-1 and CNN15-1, which were removed in CNN10-4 and CNN15-4. The results from CNN5-4, CNN10-4 and CNN15-4 were visually similar, which indicated that the cascaded CNN could compensate for the lack of depth in a single CNN. The cascaded MLP results were also presented in Figure~\ref{fig:CNNComp}, which were more noisy than the CNN results. There was slightly reduced noise level in MLP-4 compared to MLP-1, which gave MLP-4 better PSNR and SSIM than MLP-1 as shown in Figure~\ref{fig:psnr}. The coronal and sagittal view for the denoised slices were provided in Figure~\ref{fig:SagCor}, to demonstrate the uniformity across slices for the denoised results, despite of the fact that each slice were denoised independently. 

\subsection{Comparison with Conventional Denoising Methods}

9 slices from the 3 testing datasets that contained lesions were used to compare cascaded CNNs with BM3D and WNNM, and the averaged SSIM compared to the corresponding normal-dose images are listed in Table~\ref{table:SSIM}. The neural network based denoisers outperformed BM3D and WNNM for the original denoised images. After blending, BM3D and WNNM had a great gain on SSIM because the low-dose images provided details to the smooth original images. The SSIM of WNNM was higher than that of CNN5-4 and MLP-4 because it had less artifacts. The best performance were achieved by CNN15-3, which were shown in bold in Table~\ref{table:SSIM}.

\begin{table}[h]
\caption{Averaged SSIM from 9 denoisd slices}
\label{table:SSIM}
\vskip 0.15in
\begin{center}
\begin{small}
\begin{sc}
\begin{tabular}{lccr}
\hline
\abovespace\belowspace
Method & Original & Blended\\
\hline
\abovespace
Low-dose & 0.7002 & 0.7002 \\
BM3D & 0.6989 & 0.7641 \\
WNNM & 0.7163 & 0.7661 \\
CNN5-4 & 0.7611 & 0.7651 \\
CNN10-4 & 0.7624 & 0.7664 \\
CNN15-4 & \bf{0.7634} & \bf{0.7669} \\
\belowspace
MLP-4 & 0.7566 & 0.7591\\
\hline
\end{tabular}
\end{sc}
\end{small}
\end{center}
\vskip -0.1in
\end{table}

\begin{figure*}[!t]
\vskip 0.2in
\begin{center}
\centerline{\includegraphics[width=\textwidth]{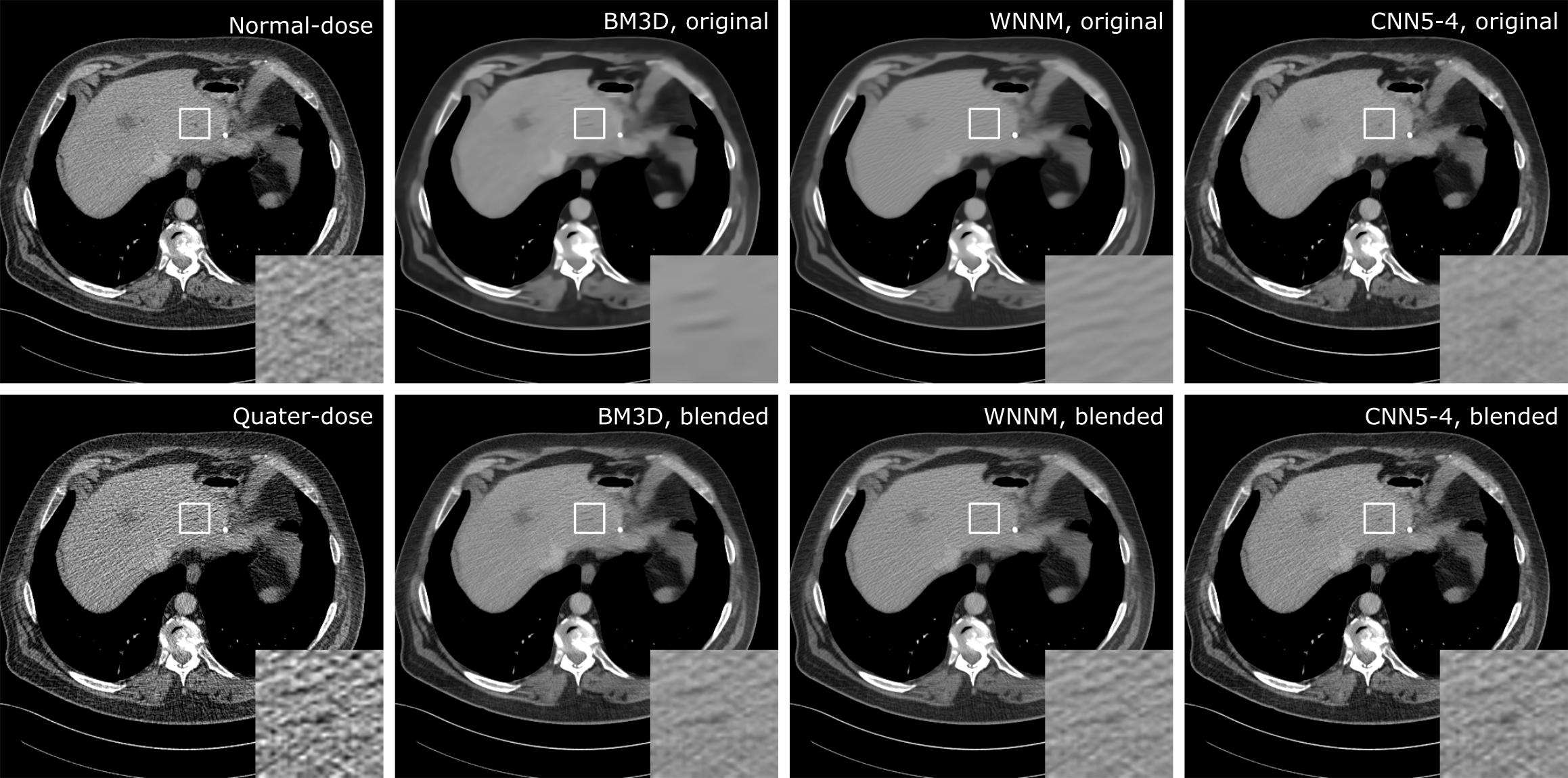}}
\caption{The denoising results with different methods on another patients' data. The area that contained a benign cyst was zoomed in and shown at the lower right corner. The grayscale window for display is $[-160\text{HU}, 240\text{HU}]$.}
\label{fig:CrossComp}
\end{center}
\vskip -0.2in
\end{figure*} 

Figure~\ref{fig:CrossComp} presented the denoising results of a benign cyst in the liver. In the original denoised results, BM3D and WNNM totally missed the lesion and only left the dark streaks near it. On the contrary, cascaded CNN was able to preserve the lesion and remove the surrounding streaks. Thus, in the final blended results, the CNN denoiser could give a more confident confirmation on the lesion than BM3D and WNNM. 

\section{Conclusion and Discussion}

In this paper, a cascaded CNN was proposed for low-dose CT image denoising. After the first CNN was trained on the training dataset, it was used to denoise the low-dose images in the training dataset. Then another CNN with the same structure was trained with the new training dataset. Cascades of CNNs could be built by repeating this scheme. The cascaded CNNs constructed a deeper neural network, and was effective at suppressing blocky and streak artifacts induced by the denoisers. The training scheme was general, the neural networks were not restricted to the CNNs used in this paper, and the application was not restricted to CT images. 

The current neural networks were implemented on 2D slices rather than 3D images. However, medical images are 3D images where the adjacent slices were highly correlated, and 3D CNNs would be able to get better results. There was no theoretical difficulty in doing that, but it requires more training data and stronger computational power. 

Although the cascaded CNN removed some of the artifacts that was generated by the CNN denoisers, it could not totally remove them and the left artifacts might cause false positivity in diagnosis. It was because some of the noises in the low-dose images resembled small lesions and CNN could not tell the difference. The best solution is to incorporate the denoising with CT image reconstructions, which can adjust the images from projection data and eliminate the artifacts generated by denoising. 

Another issue is that in this work the low-dose CT images were synthesised from normal-dose images. Although it was done with reasonable noise models, the actual noise distribution could still be different from the real situation, which could cause potential problems because the denoiser was highly adapted to a different noise model. It is nearly impossible to acquire precisely registered low- and normal-dose images from the same patient with current scanners, and mismatch in the training dataset will cause serious blurring with the current training method. Unfortunately, almost all the existing neural network based methods need to be trained with precise noise models and exactly matched training datasets. Noise model robust methods could be one of the solutions, but registration free methods also worth investigation. 

%\section*{Acknowledgements} 
% 
%\textbf{Do not} include acknowledgements in the initial version of
%the paper submitted for blind review.
%
%If a paper is accepted, the final camera-ready version can (and
%probably should) include acknowledgements. In this case, please
%place such acknowledgements in an unnumbered section at the
%end of the paper. Typically, this will include thanks to reviewers
%who gave useful comments, to colleagues who contributed to the ideas, 
%and to funding agencies and corporate sponsors that provided financial 
%support.  


\bibliography{ms}

\begin{thebibliography}{27}
\providecommand{\natexlab}[1]{#1}
\providecommand{\url}[1]{\texttt{#1}}
\expandafter\ifx\csname urlstyle\endcsname\relax
  \providecommand{\doi}[1]{doi: #1}\else
  \providecommand{\doi}{doi: \begingroup \urlstyle{rm}\Url}\fi

\bibitem[AAPM(2016)]{aapm2016challenge}
AAPM.
\newblock http://www.aapm.org/grandchallenge/lowdosect/, 2016.

\bibitem[Agostinelli et~al.(2013)Agostinelli, Anderson, and
  Lee]{agostinelli2013adaptive}
Agostinelli, F., Anderson, M.~R., and Lee, H.
\newblock Adaptive multi-column deep neural networks with application to robust
  image denoising.
\newblock In \emph{Advances in Neural Information Processing Systems}, pp.\
  1493--1501, 2013.

\bibitem[Burger et~al.(2012)Burger, Schuler, and Harmeling]{burger2012image}
Burger, H.~C., Schuler, C.~J., and Harmeling, S.
\newblock Image denoising: Can plain neural networks compete with {BM3D}?
\newblock In \emph{Computer Vision and Pattern Recognition (CVPR), 2012 IEEE
  Conference on}, pp.\  2392--2399. IEEE, 2012.

\bibitem[Chen et~al.(2017)Chen, Zhang, Kalra, Lin, Liao, Zhou, and
  Wang]{chen2017low}
Chen, H., Zhang, Y., Kalra, M.~K., Lin, F., Liao, P., Zhou, J., and Wang, G.
\newblock Low-dose {CT} with a residual encoder-decoder convolutional neural
  network {(RED-CNN)}.
\newblock \emph{arXiv preprint arXiv:1702.00288}, 2017.

\bibitem[Dabov et~al.(2007)Dabov, Foi, Katkovnik, and
  Egiazarian]{dabov2007image}
Dabov, K., Foi, A., Katkovnik, V., and Egiazarian, K.
\newblock Image denoising by sparse {3-D} transform-domain collaborative
  filtering.
\newblock \emph{IEEE Transactions on image processing}, 16\penalty0
  (8):\penalty0 2080--2095, 2007.

\bibitem[Eigen et~al.(2013)Eigen, Krishnan, and Fergus]{eigen2013restoring}
Eigen, D., Krishnan, D., and Fergus, R.
\newblock Restoring an image taken through a window covered with dirt or rain.
\newblock In \emph{Proceedings of the IEEE International Conference on Computer
  Vision}, pp.\  633--640, 2013.

\bibitem[Elbakri \& Fessler(2002)Elbakri and Fessler]{elbakri2002statistical}
Elbakri, I.~A. and Fessler, J.~A.
\newblock Statistical image reconstruction for polyenergetic {X-ray} computed
  tomography.
\newblock \emph{IEEE transactions on medical imaging}, 21\penalty0
  (2):\penalty0 89--99, 2002.

\bibitem[Fu et~al.(2016)Fu, Huang, Ding, Liao, and Paisley]{fu2016clearing}
Fu, X., Huang, J., Ding, X., Liao, Y., and Paisley, J.
\newblock Clearing the skies: A deep network architecture for single-image rain
  removal.
\newblock \emph{arXiv preprint arXiv:1609.02087}, 2016.

\bibitem[Glorot \& Bengio(2010)Glorot and Bengio]{glorot2010understanding}
Glorot, X. and Bengio, Y.
\newblock Understanding the difficulty of training deep feedforward neural
  networks.
\newblock In \emph{Aistats}, volume~9, pp.\  249--256, 2010.

\bibitem[Gu et~al.(2014)Gu, Zhang, Zuo, and Feng]{gu2014weighted}
Gu, S., Zhang, L., Zuo, W., and Feng, X.
\newblock Weighted nuclear norm minimization with application to image
  denoising.
\newblock In \emph{Proceedings of the IEEE Conference on Computer Vision and
  Pattern Recognition}, pp.\  2862--2869, 2014.

\bibitem[Han et~al.(2016)Han, Yoo, and Ye]{han2016deep}
Han, Y., Yoo, J., and Ye, J.~C.
\newblock Deep residual learning for compressed sensing {CT} reconstruction via
  persistent homology analysis.
\newblock \emph{arXiv preprint arXiv:1611.06391}, 2016.

\bibitem[He et~al.(2016)He, Zhang, Ren, and Sun]{he2016deep}
He, K., Zhang, X., Ren, S., and Sun, J.
\newblock Deep residual learning for image recognition.
\newblock In \emph{Proceedings of the IEEE Conference on Computer Vision and
  Pattern Recognition}, pp.\  770--778, 2016.

\bibitem[Hsieh(2003)]{hsieh2003computed}
Hsieh, J.
\newblock \emph{Computed tomography: principles, design, artifacts, and recent
  advances}.
\newblock SPIE press, 2003.

\bibitem[Ioffe \& Szegedy(2015)Ioffe and Szegedy]{ioffe2015batch}
Ioffe, S. and Szegedy, C.
\newblock Batch normalization: Accelerating deep network training by reducing
  internal covariate shift.
\newblock \emph{arXiv preprint arXiv:1502.03167}, 2015.

\bibitem[Jain \& Seung(2009)Jain and Seung]{jain2009natural}
Jain, V. and Seung, S.
\newblock Natural image denoising with convolutional networks.
\newblock In \emph{Advances in Neural Information Processing Systems}, pp.\
  769--776, 2009.

\bibitem[Jia et~al.(2014)Jia, Shelhamer, Donahue, Karayev, Long, Girshick,
  Guadarrama, and Darrell]{jia2014caffe}
Jia, Y., Shelhamer, E., Donahue, J., Karayev, S., Long, J., Girshick, R.,
  Guadarrama, S., and Darrell, T.
\newblock Caffe: Convolutional architecture for fast feature embedding.
\newblock In \emph{Proceedings of the 22nd ACM international conference on
  Multimedia}, pp.\  675--678. ACM, 2014.

\bibitem[Kalra et~al.(2004)Kalra, Maher, Toth, Hamberg, Blake, Shepard, and
  Saini]{kalra2004strategies}
Kalra, M.~K., Maher, M.~M., Toth, T.~L., Hamberg, L.~M., Blake, M.~A., Shepard,
  J., and Saini, S.
\newblock Strategies for {CT} radiation dose optimization 1.
\newblock \emph{Radiology}, 230\penalty0 (3):\penalty0 619--628, 2004.

\bibitem[Kang et~al.(2016)Kang, Min, and Ye]{kang2016deep}
Kang, E., Min, J., and Ye, J.~C.
\newblock A deep convolutional neural network using directional wavelets for
  low-dose {X-ray CT} reconstruction.
\newblock \emph{arXiv preprint arXiv:1610.09736}, 2016.

\bibitem[Kingma \& Ba(2014)Kingma and Ba]{kingma2014adam}
Kingma, D. and Ba, J.
\newblock Adam: A method for stochastic optimization.
\newblock \emph{arXiv preprint arXiv:1412.6980}, 2014.

\bibitem[Mao et~al.(2016)Mao, Shen, and Yang]{mao2016image}
Mao, X., Shen, C., and Yang, Y.
\newblock Image denoising using very deep fully convolutional encoder-decoder
  networks with symmetric skip connections.
\newblock \emph{arXiv preprint}, 2016.

\bibitem[Pickhardt et~al.(2012)Pickhardt, Lubner, Kim, Tang, Ruma, del Rio, and
  Chen]{pickhardt2012abdominal}
Pickhardt, P.~J., Lubner, M.~G., Kim, D.~H., Tang, J., Ruma, J.~A., del Rio,
  A.~M., and Chen, G.
\newblock Abdominal {CT} with model-based iterative reconstruction {(MBIR)}:
  initial results of a prospective trial comparing ultralow-dose with
  standard-dose imaging.
\newblock \emph{American journal of roentgenology}, 199\penalty0 (6):\penalty0
  1266--1274, 2012.

\bibitem[Prakash et~al.(2010)Prakash, Kalra, Kambadakone, Pien, Hsieh, Blake,
  and Sahani]{prakash2010reducing}
Prakash, P., Kalra, M.~K., Kambadakone, A.~K., Pien, H., Hsieh, J., Blake,
  M.~A., and Sahani, D.~V.
\newblock Reducing abdominal {CT} radiation dose with adaptive statistical
  iterative reconstruction technique.
\newblock \emph{Investigative radiology}, 45\penalty0 (4):\penalty0 202--210,
  2010.

\bibitem[Vincent et~al.(2008)Vincent, Larochelle, Bengio, and
  Manzagol]{vincent2008extracting}
Vincent, P., Larochelle, H., Bengio, Y., and Manzagol, P.
\newblock Extracting and composing robust features with denoising autoencoders.
\newblock In \emph{Proceedings of the 25th international conference on Machine
  learning}, pp.\  1096--1103. ACM, 2008.

\bibitem[Wang et~al.(2004)Wang, Bovik, Sheikh, and Simoncelli]{wang2004image}
Wang, Z., Bovik, A.~C., Sheikh, H.~R., and Simoncelli, E.~P.
\newblock Image quality assessment: from error visibility to structural
  similarity.
\newblock \emph{IEEE transactions on image processing}, 13\penalty0
  (4):\penalty0 600--612, 2004.

\bibitem[Xie et~al.(2012)Xie, Xu, and Chen]{xie2012image}
Xie, J., Xu, L., and Chen, E.
\newblock Image denoising and inpainting with deep neural networks.
\newblock In \emph{Advances in Neural Information Processing Systems}, pp.\
  341--349, 2012.

\bibitem[Xu et~al.(2014)Xu, Ren, Liu, and Jia]{xu2014deep}
Xu, L., Ren, J.~SJ., Liu, C., and Jia, J.
\newblock Deep convolutional neural network for image deconvolution.
\newblock In \emph{Advances in Neural Information Processing Systems}, pp.\
  1790--1798, 2014.

\bibitem[Zhang et~al.(2016)Zhang, Zuo, Chen, Meng, and Zhang]{zhang2016beyond}
Zhang, K., Zuo, W., Chen, Y., Meng, D., and Zhang, L.
\newblock Beyond a gaussian denoiser: Residual learning of deep {CNN} for image
  denoising.
\newblock \emph{arXiv preprint arXiv:1608.03981}, 2016.

\end{thebibliography}
\bibliographystyle{icml2017}

\end{document}